\documentclass[11pt,a4paper]{article}
\usepackage[hyperref]{acl2019}
\usepackage{times}
\usepackage{latexsym}

\usepackage{url}
\usepackage{graphicx}
\usepackage{subfigure}
\usepackage{hyperref}
\usepackage{url}
\usepackage{graphicx}
\usepackage{subfigure}
\usepackage{amssymb}
\usepackage{epsfig}
\usepackage{blindtext}
\usepackage{enumitem}
\usepackage{float}
\usepackage{amsthm}
\usepackage{wrapfig}
\usepackage{bbm}

\usepackage{amsmath}
\usepackage{booktabs}
\usepackage{multirow}
\aclfinalcopy 

\title{Depth Growing for Neural Machine Translation}

\author{Lijun Wu$^{1,*}$, Yiren Wang$^{2,}$\thanks{\;\,The first two authors contributed equally to this work. This work is conducted at Microsoft Research Asia.}, Yingce Xia$^{3,}$\thanks{\;\,Corresponding author.}, Fei Tian$^3$, Fei Gao$^3$, \\ {\bf Tao Qin$^3$, Jianhuang Lai$^1$, Tie{-}Yan Liu$^3$ }\\
  $^1$School of Data and Computer Science, Sun Yat-sen University;\\ $^2$ University of Illinois at Urbana-Champaign; $^3$ Microsoft Research Asia;\\
  $^1$\texttt{\{wulijun3, stsljh\}@mail2.sysu.edu.cn}, $^2$\texttt{yiren@illinois.edu}, \\
  $^3$\texttt{\{Yingce.Xia, fetia, feiga, taoqin, tyliu\}@microsoft.com}}

\begin{document}
\maketitle
\begin{abstract}
While very deep neural networks have shown effectiveness for computer vision and text classification applications, how to increase the network depth of neural machine translation (NMT) models for better translation quality remains a challenging problem.
Directly stacking more blocks to the NMT model results in no improvement and even reduces performance. In this work, we propose an effective two-stage approach with three specially designed components to construct deeper NMT models, which results in significant improvements over the strong Transformer baselines on WMT$14$ English$\to$German and English$\to$French translation tasks\footnote{{Our code is available at \url{https://github.com/apeterswu/Depth_Growing_NMT}}}.
\end{abstract}

\section{Introduction}

Neural machine translation (briefly, NMT), which is built upon deep neural networks, has gained rapid progress in recent years~\citep{bahdanau2015neural,sutskever2014sequence,sennrich2016neural,he2016dual,sennrich2016improving,xia2017deliberation,wang2018multiagent} and achieved significant improvement in translation quality~\citep{hassan2018achieving}. Variants of network structures have been applied in NMT such as LSTM~\citep{wu2016google}, CNN~\citep{gehring2017convolutional} and Transformer~\citep{vaswani2017attention}.

\begin{figure}[!t]
\centering
\includegraphics[width=1.0\linewidth]{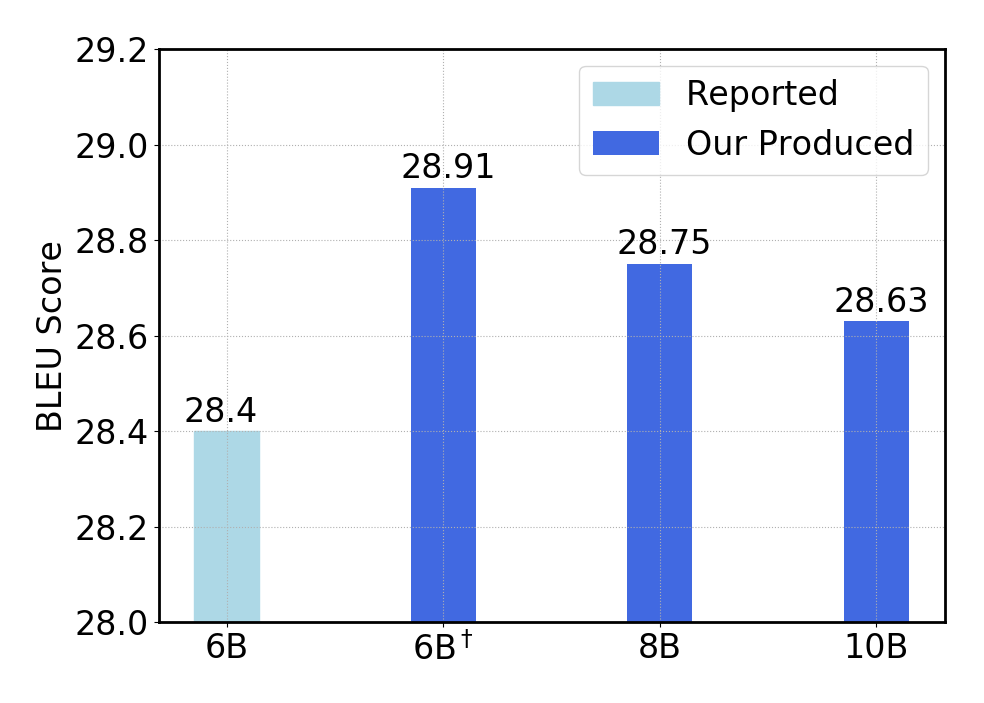}
\caption{Performances of Transformer models with different number of encoder/decoder blocks (recorded on $x$-axis) on WMT$14$ En$\to$De translation task. $\dagger$ denotes the result reported in \cite{vaswani2017attention}.}
\label{fig:baselines}
\end{figure}

Training deep networks has always been a challenging problem, mainly due to the difficulties in optimization for deep architecture. Breakthroughs have been made in computer vision to enable deeper model construction via advanced initialization schemes~\citep{he2015delving}, multi-stage training strategy~\citep{simonyan2015very}, and novel model architectures~\citep{srivastava2015highway,he2016deep}. While constructing very deep neural networks with tens and even more than a hundred blocks have shown effectiveness in image recognition~\citep{he2016deep}, question answering and text classification~\citep{devlin2018bert,radford2019language}, scaling up model capacity with very deep network remains challenging for NMT. The NMT models are generally constructed with up to $6$ encoder and decoder blocks in both state-of-the-art research work and champion systems of machine translation competition. For example, the LSTM-based models are usually stacked for $4$~\citep{stahlberg2018university} or $6$~\citep{chen2018best} blocks, and the state-of-the-art Transformer models are equipped with a $6$-block encoder and decoder~\citep{vaswani2017attention, junczys2018microsoft,edunov2018understanding}. Increasing the NMT model depth by directly stacking more blocks results in no improvement or performance drop (Figure~\ref{fig:baselines}), and even leads to optimization failure~\citep{bapna2018training}. 

There have been a few attempts in previous works on constructing deeper NMT models. \citet{zhou2016deep} and \citet{wang2017deep} propose increasing the depth of LSTM-based models by introducing linear units between internal hidden states to eliminate the problem of gradient vanishing. However, their methods are specially designed for the recurrent architecture which has been significantly outperformed by the state-of-the-art transformer model. \citet{bapna2018training} propose an enhancement to the attention mechanism to ease the optimization of models with deeper encoders. While gains have been reported over different model architectures including LSTM and Transformer, their improvements are not made over the best performed baseline model configuration. How to construct and train {\em deep} NMT models to push forward the state-of-the-art translation performance with larger model capacity remains a challenging and open problem.

In this work, we explore the potential of leveraging {\em deep} neural networks for NMT and propose a new approach to construct and train deeper NMT models. As aforementioned, constructing deeper models is not as straightforward as directly stacking more blocks, but requires new mechanisms to boost the training and utilize the larger capacity with minimal increase in complexity. Our solution is a new two-stage training strategy, which ``grows'' a well-trained NMT model into a deeper network with three components specially designed to overcome the optimization difficulty and best leverage the capability of both shallow and deep architecture. Our approach can effectively construct a deeper model with significantly better performance, and is generally applicable to any model architecture.

We evaluate our approach on two  large-scale benchmark datasets, WMT$14$ English$\to$German and English$\to$French translations. Empirical studies show that our approach can significantly improve in translation quality with an increased model depth. Specifically, we achieve $1.0$ and $0.6$ BLEU score improvement over the strong Transformer baseline in English$\to$German and English$\to$French translations.

\begin{figure}[!t]
\centering
\includegraphics[width=0.9\linewidth]{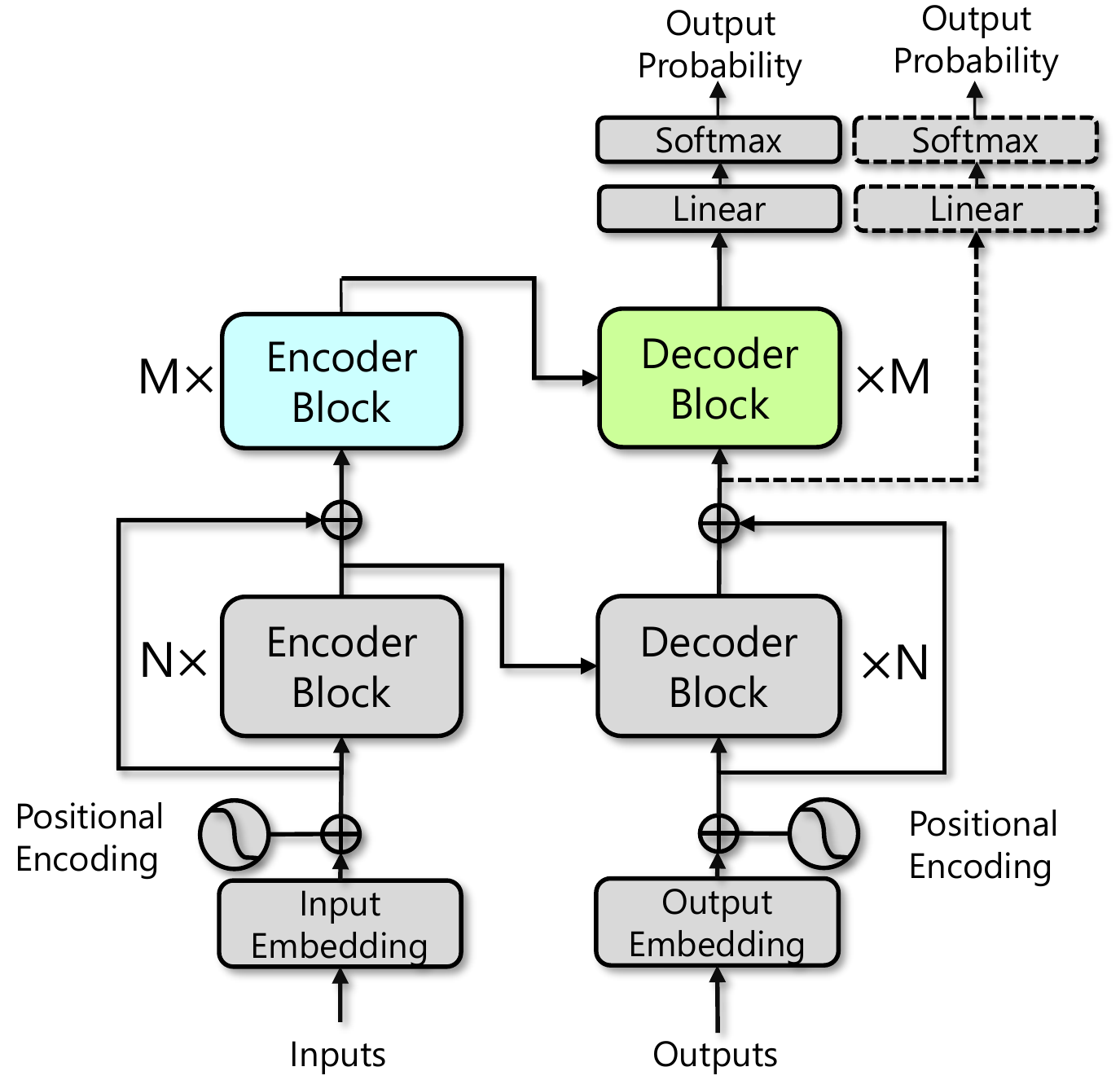}
\caption{The overall framework of our proposed deep model architecture. $N$ and $M$ are the numbers of blocks in the bottom module (i.e., grey parts) and top module (i.e., blue and green parts). Parameters of the bottom module are fixed during the top module training. The dashed parts denote the original training/decoding of the bottom module. The weights of the two linear operators before softmax are shared.}
\label{fig:arch}
\end{figure}

\section{Approach}
\label{sec:approach}
We introduce the details of our proposed approach in this section. The overall framework is illustrated in Figure~\ref{fig:arch}. 

Our model consists of a bottom module with $N$ blocks of encoder and decoder (the grey components in Figure~\ref{fig:arch}), and a top module with $M$ blocks (the blue and green components). We denote the encoder and decoder of the bottom module as $\texttt{enc}_1$ and $\texttt{dec}_1$, and the corresponding two parts of the top module as $\texttt{enc}_2$ and $\texttt{dec}_2$. An encoder-decoder attention mechanism is used in the decoder blocks of the NMT models, and here we use $\texttt{attn}_1$ and $\texttt{attn}_2$ to represent such attention in the bottom and top modules respectively. 

The model is constructed via a two-stage training strategy: in Stage 1, the bottom module (i.e., $\texttt{enc}_1$ and $\texttt{dec}_1$) is trained and subsequently holds constant; in Stage 2,  only the top module (i.e., $\texttt{enc}_2$ and $\texttt{dec}_2$) is optimized. 

Let $x$ and $y$ denote the embedding of source and target sequence. Let $l_y$ denote the number of words in $y$, and $y_{<t}$ denote the elements before time step $t$. Our proposed model works in the following way:
\begin{align}
& h_1 =\texttt{enc}_1(x);\;h_2=\texttt{enc}_2(x+h_1);\label{eq:method:encoder}\\
& s_{1,t} = \texttt{dec}_1(y_{<t},\texttt{attn}_1(h_1)),\;\forall t\in[l_y];\label{eq:method:Dec1} \\
& s_{2,t}=\texttt{dec}_2(y_{<t}+s_{1,<t},\texttt{attn}_2(h_2))\label{eq:method:Dec2},
\end{align}
which contains three key components specially designed for deeper model construction, including:

\noindent(1) {\bf{\em Cross-module residual connections:}} As shown in Eqn.\eqref{eq:method:encoder}, the encoder $\texttt{enc}_1$ of the bottom module encodes the input $x$ to a hidden representation $h_1$, then a cross-module residual connection is introduced to the top module and the representation $h_2$ is eventually produced. The decoders work in a similar way as shown in Eqn.\eqref{eq:method:Dec1} and \eqref{eq:method:Dec2}. This enables the top module to have direct access to both the low-level input signals from the word embedding and high-level information generated by the bottom module. Similar principles can be found in ~\citet{wang2017deep,wu2018word}.

\noindent(2) {\bf{\em Hierarchical encoder-decoder attention:}} We introduce a hierarchical encoder-decoder attention calculated with different contextual representations as shown in Eqn.\eqref{eq:method:Dec1} and \eqref{eq:method:Dec2}, where $h_1$ is used as key and value for $\texttt{attn}_1$ in the bottom module, and $h_2$ for $\texttt{attn}_2$ in the top module. Hidden states from the corresponding previous decoder block are used as queries for both $\texttt{attn}_1$ and $\texttt{attn}_2$ (omitted for readability). In this way, the strong capability of the well trained bottom module can be best preserved regardless of the influence from top module, while the newly stacked top module can leverage the higher-level contextual representations. More details can be found from source code in the supplementary materials.

\noindent(3) {\bf{\em Deep-shallow decoding:}} At the decoding phase, $\texttt{enc}_1$ and $\texttt{dec}_1$ work together according to Eqn.\eqref{eq:method:encoder} and Eqn.\eqref{eq:method:Dec1} as a shallow network $\texttt{net}_S$, integrate both bottom and top module works as a deep network $\texttt{net}_D$ according to Eqn.\eqref{eq:method:encoder}$\sim$Eqn.\eqref{eq:method:Dec2}. $\texttt{net}_S$ and $\texttt{net}_D$ generate the final translation results through reranking.


\noindent {\bf Discussion}

\noindent $\bullet$ {\em Training complexity:} As aforementioned, the bottom module is trained in Stage 1 and only parameters of the top module are optimized in Stage 2. This significantly eases optimization difficulty and reduces training complexity. Jointly training the two modules with minimal training complexity is left for future work.

\noindent $\bullet$ {\em Ensemble learning:} What we propose in this paper is a {\em single} deeper model with hierarchical contextual information, although the deep-shallow decoding is similar to the ensemble methods in terms of inference complexity~\citep{zhou2012ensemble}. While training multiple diverse models for good ensemble performance introduces high additional complexity, our approach, as discussed above, ``grows" a well-trained model into a deeper one with minimal increase in training complexity. Detailed empirical analysis is presented in Section~\ref{subsec:analysis}.

\section{Experiments}
We evaluate our proposed approach on two large-scale benchmark datasets. We compare our approach with multiple baseline models, and analyze the effectiveness of our deep training strategy.

\subsection{Experiment Design}
\paragraph{Datasets}
We conduct experiments to evaluate the effectiveness of our proposed method on two widely adopted benchmark datasets: the WMT$14$\footnote{\url{http://www.statmt.org/wmt14/translation-task.html}} English$\to$German translation (En$\to$De) and the WMT$14$ English$\to$French translation (En$\to$Fr). We use $4.5M$ parallel sentence pairs for En$\to$De and $36M$ pairs for En$\to$Fr as our training data\footnote{Training data are constructed with filtration rules following \url{https://github.com/pytorch/fairseq/tree/master/examples/translation}}. We use the concatenation of \emph{Newstest2012} and \emph{Newstest2013} as the validation set, and \emph{Newstest2014} as the test set. All words are segmented into sub-word units using byte pair encoding (BPE)\footnote{\url{https://github.com/rsennrich/subword-nmt}}~\citep{sennrich2016neural}, forming a vocabulary shared by the source and target languages with $32k$ and $45k$ tokens for En$\to$De and En$\to$Fr respectively.

\paragraph{Architecture}
The basic encoder-decoder framework we use is the strong Transformer model. We adopt the \texttt{big} transformer configuration following~\citet{vaswani2017attention}, with the dimension of word embeddings, hidden states and non-linear layer set as $1024$, $1024$ and $4096$ respectively. The dropout rate is $0.3$ for En$\to$De and $0.1$ for En$\to$Fr. We set the number of encoder/decoder blocks for the bottom module as $N=6$ following the common practice, and set the number of additionally stacked blocks of the top module as $M=2$. Our models are implemented based on the PyTorch implementation of Transformer\footnote{\url{https://github.com/pytorch/fairseq}} and the code can be found in the supplementary materials.

\paragraph{Training}
We use Adam~\citep{kingma2015adam} optimizer following the optimization settings and default learning rate schedule in~\citet{vaswani2017attention} for model training. All models are trained on $8$ M40 GPUs.

\paragraph{Evaluation}
We evaluate the model performances with tokenized case-sensitive BLEU\footnote{\url{https://github.com/moses-smt/mosesdecoder/blob/master/scripts/generic/multi-bleu.perl}} score~\citep{papineni2002bleu} 
for the two translation tasks. We use beam search with a beam size of $5$ and with no length penalty. 

\subsection{Overall Results}
We compare our method ({\em Ours}) with the Transformer baselines of $6$ blocks ($6B$) and $8$ blocks ($8B$), and a 16-block Transformer with transparent attention ({\em Transparent Attn (16B)})\footnote{We directly use the performance figure from~\citep{bapna2018training}, which uses the \texttt{base} Transformer configuration. We run the method of our own implementation with the widely adopted and state-of-the-art \texttt{big} setting, but no improvement has been observed.}~\citep{bapna2018training}. We also reproduce a $6$-block Transformer baseline, which has better performance than what is reported in~\cite{vaswani2017attention} and we use it to initialize the bottom module in our model.

\begin{table}[!tb]
\centering
\caption{The test set performances of WMT$14$ En$\to$De and En$\to$Fr translation tasks. `$\dagger$' denotes the performance figures reported in the previous works.}
\begin{tabular}{lll}
\toprule
Model                               & En$\to$De    & En$\to$Fr    \\ \midrule
Transformer (6B)$^\dagger$          & $28.40$      & $41.80$      \\
Transformer (6B)                    & $28.91$      & $42.69$      \\
Transformer (8B)                    & $28.75$      & $42.63$      \\
Transformer (10B)                   & $28.63$      & $42.73$      \\
Transparent Attn (16B)$^\dagger$    & $28.04$      & $-$          \\
\bf{Ours (8B)}                      & $\bf{29.92}$ & $\bf{43.27}$ \\
\bottomrule
\end{tabular}
\label{tab:results-wmt}
\end{table}

From the results in Table~\ref{tab:results-wmt}, we see that our proposed approach enables effective training for deeper network and achieves  significantly better performances compared to baselines. With our method, the performance of a well-optimized $6$-block model can be further boosted by adding two additional blocks, while simply using Transformer (8B) will lead to a performance drop. Specifically, we achieve a $29.92$ BLEU score on En$\to$De translation with $1.0$ BLEU improvement over the strong baselines, and achieve a $0.6$ BLEU improvement for En$\to$Fr. The improvements are statistically significant with $p < 0.01$ in paired bootstrap sampling~\citep{koehn2004statistical}. 

We further make an attempt to train a deeper model with additional $M=4$ blocks, which has $10$ blocks in total for En$\rightarrow$De translation. The bottom module is also initialized from our reproduced $6$-block transformer baseline. This model achieves a $30.07$ BLEU score on En$\rightarrow$De translation and it surpasses the performance of our $8$-block model, which further demonstrates that our approach is effective for training deeper NMT models.

\subsection{Analysis}
\label{subsec:analysis}
To further study the effectiveness of our proposed framework, we present additional comparisons in En$\to$De translation with two groups of baseline approaches in Figure~\ref{fig:results-study}:

\begin{figure}[!tb]
    \centering
    \includegraphics[width=1.0\linewidth]{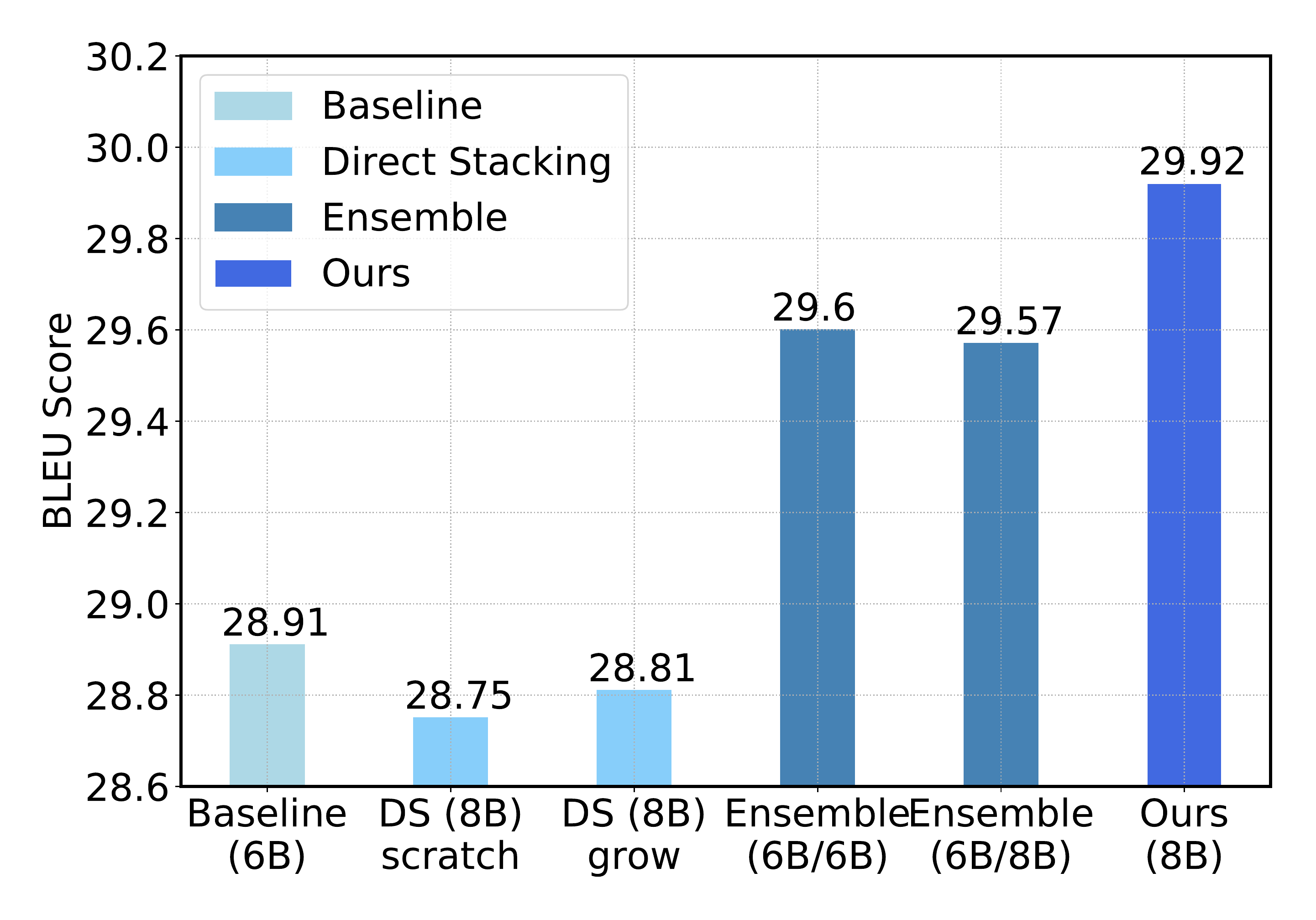}
    \caption{The test performances of WMT$14$ En$\to$De translation task.}
    \label{fig:results-study}
\end{figure}

\noindent (1) Direct stacking ({\em DS}): we extend the $6$-block baseline to $8$-block by directly stacking $2$ additional blocks. 
We can see that both training from scratch ({\em DS scratch}) and ``growing'' from a well-trained $6$-block model ({\em DS grow}) fails to improve performance in spite of larger model capacity. The comparison with this group of models shows that directly stacking more blocks is not a good strategy for increasing network depth, and demonstrates the effectiveness and necessity of our proposed mechanisms for training deep networks.

\noindent (2) Ensemble learning ({\em Ensemble}): we present the two-model ensemble results for fair comparison with our approach that involves a two-pass deep-shallow decoding. Specifically, we present the ensemble performances of two independently trained $6$-block models ({\em Ensemble $6$B/$6$B}), and ensemble of one $6$-block and one $8$-block model independently trained from scratch ({\em Ensemble $6$B/$8$B}). As expected, the ensemble method improves translation quality over the single model baselines by a large margin (over $0.8$ BLEU improvement). Regarding training complexity, it takes $40$ GPU days ($5$ days on $8$ GPU) to train a single $6$-block model from scratch, $48$ GPU days for a $8$-block model , and $8$ GPU days to ``grow'' a $6$-block model into $8$-block with our approach. Therefore, our model is better than the two-model ensemble in terms of both translation quality (more than $0.3$ BLEU improvement over the ensemble baseline) and training complexity. 

\section{Conclusion}
In this paper, we proposed a new training strategy with three specially designed components, including cross-module residual connection, hierarchical encoder-decoder attention and deep-shallow decoding, to construct and train deep NMT models. We showed that our approach can effectively construct deeper model with significantly better performance over the state-of-the-art transformer baseline. Although only empirical studies on the transformer are presented in this paper, our proposed strategy is a general approach that can be universally applicable to other model architectures, including LSTM and CNN. In future work, we will further explore efficient strategies that can jointly train all modules of the deep model with minimal increase in training complexity.

\bibliographystyle{acl_natbib}
\bibliography{acl2019}

\begin{thebibliography}{29}
\expandafter\ifx\csname natexlab\endcsname\relax\def\natexlab#1{#1}\fi

\bibitem[{Bahdanau et~al.(2015)Bahdanau, Cho, and Bengio}]{bahdanau2015neural}
Dzmitry Bahdanau, Kyunghyun Cho, and Yoshua Bengio. 2015.
\newblock Neural machine translation by jointly learning to align and
  translate.
\newblock In \emph{Third International Conference on Learning Representations}.

\bibitem[{Bapna et~al.(2018)Bapna, Chen, Firat, Cao, and
  Wu}]{bapna2018training}
Ankur Bapna, Mia Chen, Orhan Firat, Yuan Cao, and Yonghui Wu. 2018.
\newblock Training deeper neural machine translation models with transparent
  attention.
\newblock In \emph{Proceedings of the 2018 Conference on Empirical Methods in
  Natural Language Processing}, pages 3028--3033.

\bibitem[{Chen et~al.(2018)Chen, Firat, Bapna, Johnson, Macherey, Foster,
  Jones, Schuster, Shazeer, Parmar et~al.}]{chen2018best}
Mia~Xu Chen, Orhan Firat, Ankur Bapna, Melvin Johnson, Wolfgang Macherey,
  George Foster, Llion Jones, Mike Schuster, Noam Shazeer, Niki Parmar, et~al.
  2018.
\newblock The best of both worlds: Combining recent advances in neural machine
  translation.
\newblock In \emph{Proceedings of the 56th Annual Meeting of the Association
  for Computational Linguistics (Volume 1: Long Papers)}, pages 76--86.

\bibitem[{Devlin et~al.(2018)Devlin, Chang, Lee, and
  Toutanova}]{devlin2018bert}
Jacob Devlin, Ming-Wei Chang, Kenton Lee, and Kristina Toutanova. 2018.
\newblock Bert: Pre-training of deep bidirectional transformers for language
  understanding.
\newblock \emph{arXiv preprint arXiv:1810.04805}.

\bibitem[{Edunov et~al.(2018)Edunov, Ott, Auli, and
  Grangier}]{edunov2018understanding}
Sergey Edunov, Myle Ott, Michael Auli, and David Grangier. 2018.
\newblock Understanding back-translation at scale.
\newblock In \emph{Proceedings of the 2018 Conference on Empirical Methods in
  Natural Language Processing}, pages 489--500.

\bibitem[{Gehring et~al.(2017)Gehring, Auli, Grangier, Yarats, and
  Dauphin}]{gehring2017convolutional}
Jonas Gehring, Michael Auli, David Grangier, Denis Yarats, and Yann~N Dauphin.
  2017.
\newblock Convolutional sequence to sequence learning.
\newblock In \emph{Proceedings of the 34th International Conference on Machine
  Learning-Volume 70}, pages 1243--1252. JMLR. org.

\bibitem[{Hassan et~al.(2018)Hassan, Aue, Chen, Chowdhary, Clark, Federmann,
  Huang, Junczys-Dowmunt, Lewis, Li et~al.}]{hassan2018achieving}
Hany Hassan, Anthony Aue, Chang Chen, Vishal Chowdhary, Jonathan Clark,
  Christian Federmann, Xuedong Huang, Marcin Junczys-Dowmunt, William Lewis,
  Mu~Li, et~al. 2018.
\newblock Achieving human parity on automatic chinese to english news
  translation.
\newblock \emph{arXiv preprint arXiv:1803.05567}.

\bibitem[{He et~al.(2016{\natexlab{a}})He, Xia, Qin, Wang, Yu, Liu, and
  Ma}]{he2016dual}
Di~He, Yingce Xia, Tao Qin, Liwei Wang, Nenghai Yu, Tie-Yan Liu, and Wei-Ying
  Ma. 2016{\natexlab{a}}.
\newblock Dual learning for machine translation.
\newblock In \emph{Advances in Neural Information Processing Systems}, pages
  820--828.

\bibitem[{He et~al.(2015)He, Zhang, Ren, and Sun}]{he2015delving}
Kaiming He, Xiangyu Zhang, Shaoqing Ren, and Jian Sun. 2015.
\newblock Delving deep into rectifiers: Surpassing human-level performance on
  imagenet classification.
\newblock In \emph{Proceedings of the IEEE international conference on computer
  vision}, pages 1026--1034.

\bibitem[{He et~al.(2016{\natexlab{b}})He, Zhang, Ren, and Sun}]{he2016deep}
Kaiming He, Xiangyu Zhang, Shaoqing Ren, and Jian Sun. 2016{\natexlab{b}}.
\newblock Deep residual learning for image recognition.
\newblock In \emph{Proceedings of the IEEE conference on computer vision and
  pattern recognition}, pages 770--778.

\bibitem[{Junczys-Dowmunt(2018)}]{junczys2018microsoft}
Marcin Junczys-Dowmunt. 2018.
\newblock Microsoft's submission to the wmt2018 news translation task: How i
  learned to stop worrying and love the data.
\newblock In \emph{Proceedings of the Third Conference on Machine Translation:
  Shared Task Papers}, pages 425--430.

\bibitem[{Kingma and Ba(2015)}]{kingma2015adam}
Diederik~P Kingma and Jimmy Ba. 2015.
\newblock Adam: A method for stochastic optimization.
\newblock In \emph{Third International Conference on Learning Representations}.

\bibitem[{Koehn(2004)}]{koehn2004statistical}
Philipp Koehn. 2004.
\newblock Statistical significance tests for machine translation evaluation.
\newblock In \emph{Proceedings of the 2004 conference on empirical methods in
  natural language processing}.

\bibitem[{Papineni et~al.(2002)Papineni, Roukos, Ward, and
  Zhu}]{papineni2002bleu}
Kishore Papineni, Salim Roukos, Todd Ward, and Wei-Jing Zhu. 2002.
\newblock Bleu: a method for automatic evaluation of machine translation.
\newblock In \emph{Proceedings of the 40th annual meeting on association for
  computational linguistics}, pages 311--318. Association for Computational
  Linguistics.

\bibitem[{Radford et~al.(2019)Radford, Wu, Child, Luan, Amodei, and
  Sutskever}]{radford2019language}
Alec Radford, Jeff Wu, Rewon Child, David Luan, Dario Amodei, and Ilya
  Sutskever. 2019.
\newblock Language models are unsupervised multitask learners.

\bibitem[{Sennrich et~al.(2016{\natexlab{a}})Sennrich, Haddow, and
  Birch}]{sennrich2016improving}
Rico Sennrich, Barry Haddow, and Alexandra Birch. 2016{\natexlab{a}}.
\newblock Improving neural machine translation models with monolingual data.
\newblock In \emph{Proceedings of the 54th Annual Meeting of the Association
  for Computational Linguistics (Volume 1: Long Papers)}, volume~1, pages
  86--96.

\bibitem[{Sennrich et~al.(2016{\natexlab{b}})Sennrich, Haddow, and
  Birch}]{sennrich2016neural}
Rico Sennrich, Barry Haddow, and Alexandra Birch. 2016{\natexlab{b}}.
\newblock Neural machine translation of rare words with subword units.
\newblock In \emph{Proceedings of the 54th Annual Meeting of the Association
  for Computational Linguistics (Volume 1: Long Papers)}, volume~1, pages
  1715--1725.

\bibitem[{Simonyan and Zisserman(2015)}]{simonyan2015very}
Karen Simonyan and Andrew Zisserman. 2015.
\newblock Very deep convolutional networks for large-scale image recognition.
\newblock In \emph{Third International Conference on Learning Representations}.

\bibitem[{Srivastava et~al.(2015)Srivastava, Greff, and
  Schmidhuber}]{srivastava2015highway}
Rupesh~Kumar Srivastava, Klaus Greff, and J{\"u}rgen Schmidhuber. 2015.
\newblock Highway networks.
\newblock \emph{arXiv preprint arXiv:1505.00387}.

\bibitem[{Stahlberg et~al.(2018)Stahlberg, de~Gispert, and
  Byrne}]{stahlberg2018university}
Felix Stahlberg, Adri{\`a} de~Gispert, and Bill Byrne. 2018.
\newblock The university of cambridge’s machine translation systems for
  wmt18.
\newblock In \emph{Proceedings of the Third Conference on Machine Translation:
  Shared Task Papers}, pages 504--512.

\bibitem[{Sutskever et~al.(2014)Sutskever, Vinyals, and
  Le}]{sutskever2014sequence}
Ilya Sutskever, Oriol Vinyals, and Quoc~V Le. 2014.
\newblock Sequence to sequence learning with neural networks.
\newblock In \emph{Advances in neural information processing systems}, pages
  3104--3112.

\bibitem[{Vaswani et~al.(2017)Vaswani, Shazeer, Parmar, Uszkoreit, Jones,
  Gomez, Kaiser, and Polosukhin}]{vaswani2017attention}
Ashish Vaswani, Noam Shazeer, Niki Parmar, Jakob Uszkoreit, Llion Jones,
  Aidan~N Gomez, {\L}ukasz Kaiser, and Illia Polosukhin. 2017.
\newblock Attention is all you need.
\newblock In \emph{Advances in Neural Information Processing Systems}, pages
  5998--6008.

\bibitem[{Wang et~al.(2017)Wang, Lu, Zhou, and Liu}]{wang2017deep}
Mingxuan Wang, Zhengdong Lu, Jie Zhou, and Qun Liu. 2017.
\newblock Deep neural machine translation with linear associative unit.
\newblock In \emph{Proceedings of the 55th Annual Meeting of the Association
  for Computational Linguistics (Volume 1: Long Papers)}, volume~1, pages
  136--145.

\bibitem[{Wang et~al.(2019)Wang, Xia, He, Tian, Qin, Zhai, and
  Liu}]{wang2018multiagent}
Yiren Wang, Yingce Xia, Tianyu He, Fei Tian, Tao Qin, ChengXiang Zhai, and
  Tie-Yan Liu. 2019.
\newblock \href {https://openreview.net/forum?id=HyGhN2A5tm} {Multi-agent dual
  learning}.
\newblock In \emph{Seventh International Conference on Learning
  Representations}.

\bibitem[{Wu et~al.(2018)Wu, Tian, Zhao, Lai, and Liu}]{wu2018word}
Lijun Wu, Fei Tian, Li~Zhao, Jianhuang Lai, and Tie-Yan Liu. 2018.
\newblock Word attention for sequence to sequence text understanding.
\newblock In \emph{Thirty-Second AAAI Conference on Artificial Intelligence}.

\bibitem[{Wu et~al.(2016)Wu, Schuster, Chen, Le, Norouzi, Macherey, Krikun,
  Cao, Gao, Macherey et~al.}]{wu2016google}
Yonghui Wu, Mike Schuster, Zhifeng Chen, Quoc~V Le, Mohammad Norouzi, Wolfgang
  Macherey, Maxim Krikun, Yuan Cao, Qin Gao, Klaus Macherey, et~al. 2016.
\newblock Google's neural machine translation system: Bridging the gap between
  human and machine translation.
\newblock \emph{arXiv preprint arXiv:1609.08144}.

\bibitem[{Xia et~al.(2017)Xia, Tian, Wu, Lin, Qin, Yu, and
  Liu}]{xia2017deliberation}
Yingce Xia, Fei Tian, Lijun Wu, Jianxin Lin, Tao Qin, Nenghai Yu, and Tie-Yan
  Liu. 2017.
\newblock Deliberation networks: Sequence generation beyond one-pass decoding.
\newblock In \emph{Advances in Neural Information Processing Systems}, pages
  1784--1794.

\bibitem[{Zhou et~al.(2016)Zhou, Cao, Wang, Li, and Xu}]{zhou2016deep}
Jie Zhou, Ying Cao, Xuguang Wang, Peng Li, and Wei Xu. 2016.
\newblock Deep recurrent models with fast-forward connections for neural
  machine translation.
\newblock \emph{Transactions of the Association for Computational Linguistics},
  4:371--383.

\bibitem[{Zhou(2012)}]{zhou2012ensemble}
Zhi-Hua Zhou. 2012.
\newblock \emph{Ensemble methods: foundations and algorithms}.
\newblock Chapman and Hall/CRC.

\end{thebibliography}

\end{document}